\documentclass[conference]{IEEEtran}
\IEEEoverridecommandlockouts
\usepackage{cite}
\usepackage{amsmath,amssymb,amsfonts}
\usepackage{algorithmic}
\usepackage{graphicx}
\usepackage{textcomp}
\usepackage{xcolor}
\usepackage{url}
\usepackage{subcaption}
\usepackage{multirow}
\usepackage{siunitx}
\usepackage{etoolbox}
\newcommand{\rulesep}{\unskip\ \vrule\ }

\def\BibTeX{{\rm B\kern-.05em{\sc i\kern-.025em b}\kern-.08em
    T\kern-.1667em\lower.7ex\hbox{E}\kern-.125emX}}
\begin{document}

\title{Semantic SuperPoint: A Deep Semantic Descriptor\\
\thanks{This work was supported by the São Paulo Research
Foundation - FAPESP (grant number 2021/08117-0 and
2014/50851-0), by the Brazilian National Council for
Scientific and Technological Development (grant number
465755/2014-3), and by the Coordination of Improvement
of Higher Education Personnel - Brazil (Finance Code 001). 

%Code available at: \url{https://github.com/Gabriel-SGama/Semantic-SuperPoint}
Code available at: https://github.com/Gabriel-SGama/Semantic-SuperPoint
}}
%\author{\IEEEauthorblockN{Gabriel S. Gama, Nícolas S. Rosa and Valdir Grassi Jr.}
%\IEEEauthorblockA{São Carlos School of Engineering\\
%University of São Paulo\\
%São Carlos, SP, Brazil\\
%Email: \{gabriel\_gama,nicolas.rosa,vgrassi\}@usp.br}}

 \author{\IEEEauthorblockN{Gabriel Soares Gama}
 \IEEEauthorblockA{\textit{São Carlos School of Engineering} \\
 \textit{University of São Paulo}\\
 São Carlos, SP, Brazil \\
 \texttt{gabriel\_gama@usp.br}
 % \orcidlink{0000-0002-4866-5197}
 }
 \and
 \IEEEauthorblockN{Nícolas dos Santos Rosa}
 \IEEEauthorblockA{\textit{São Carlos School of Engineering} \\
 \textit{University of São Paulo}\\
 São Carlos, SP, Brazil \\
 \texttt{nicolas.rosa@usp.br}}
 \and
 \IEEEauthorblockN{Valdir Grassi Jr.}
 \IEEEauthorblockA{\textit{São Carlos School of Engineering} \\
 \textit{University of São Paulo}\\
 São Carlos, SP, Brazil \\
 % email address or ORCID
 \texttt{vgrassi@usp.br}}}

 % \texttt{gabriel\_gama@usp.br}, \texttt{nicolas.rosa@usp.br} and \texttt{vgrassi@usp.br}

\maketitle

\begin{abstract}
Several SLAM methods benefit from the use of semantic information. Most integrate photometric methods with high-level semantics such as object detection and semantic segmentation. We propose that adding a semantic segmentation decoder in a shared encoder architecture would help the descriptor decoder learn semantic information, improving the feature extractor. This would be a more robust approach than only using high-level semantic information since it would be intrinsically learned in the descriptor and would not depend on the final quality of the semantic prediction. To add this information, we take advantage of multi-task learning methods to improve accuracy and balance the performance of each task. The proposed models are evaluated according to detection and matching metrics on the HPatches dataset. The results show that the Semantic SuperPoint model performs better than the baseline one.

\end{abstract}

\begin{IEEEkeywords}
Computer Vision, Visual Odometry
\end{IEEEkeywords}

\section{Introduction}

Visual Simultaneous Localization and Mapping (vSLAM) is the process of creating a map based on the visual information without a global reference, being used in multiple applications where GPS signal is not available or precise enough. In the context of vSLAM based on features, this is done by using the relative pose obtained by detecting interest points, generating a descriptor for them and matching the extracted keypoint in the previous frame to the current one. Those matches are then processed using algorithms similar to the normalized eight-point algorithm \cite{8point_alg}. On top of that, loop closing and global optimization processes are normally implemented to improve the estimated trajectory, removing drift error. 

All of that is dependent on the correct association of keypoints. Most feature extraction methods generate descriptors based solely on photometric information like ORB \cite{ORB}, SURF \cite{SURF} and SIFT \cite{SIFT}, with only the first one being both fast and reliable. Although they achieve excellent results in SLAM systems, handmade descriptors are highly susceptible to illumination changes, which decreases the quality of the pose estimation. This problem is important to be considered since most SLAM applications are made to run in an uncontrolled environment. 

Convolutional neural networks are normally used in computer vision tasks to increase robustness, and most often obtain better results than handmade methods \cite{conv_better_hand1, conv_better_hand2}. Similarly, methods such as SuperPoint \cite{superpoint} and LIFT \cite{LIFT} were developed to improve the coherency of feature extraction algorithms. Both are superior to ORB and present comparable or better results than SIFT while keeping a real-time performance.

In the context of deep learning, models that learn multiple related tasks can have a better result if trained with a shared encoder due to the inductive bias, as seen in the related literature \cite{MTL_kendall, MTL_ang}, \cite{MTL_sener}. In comparison to using multiple models and soft parameter sharing, the computational cost of using a shared encoder for multi-task is relatively low.

\begin{figure}[t]
    \centering
    \includegraphics[width=.4\textwidth]{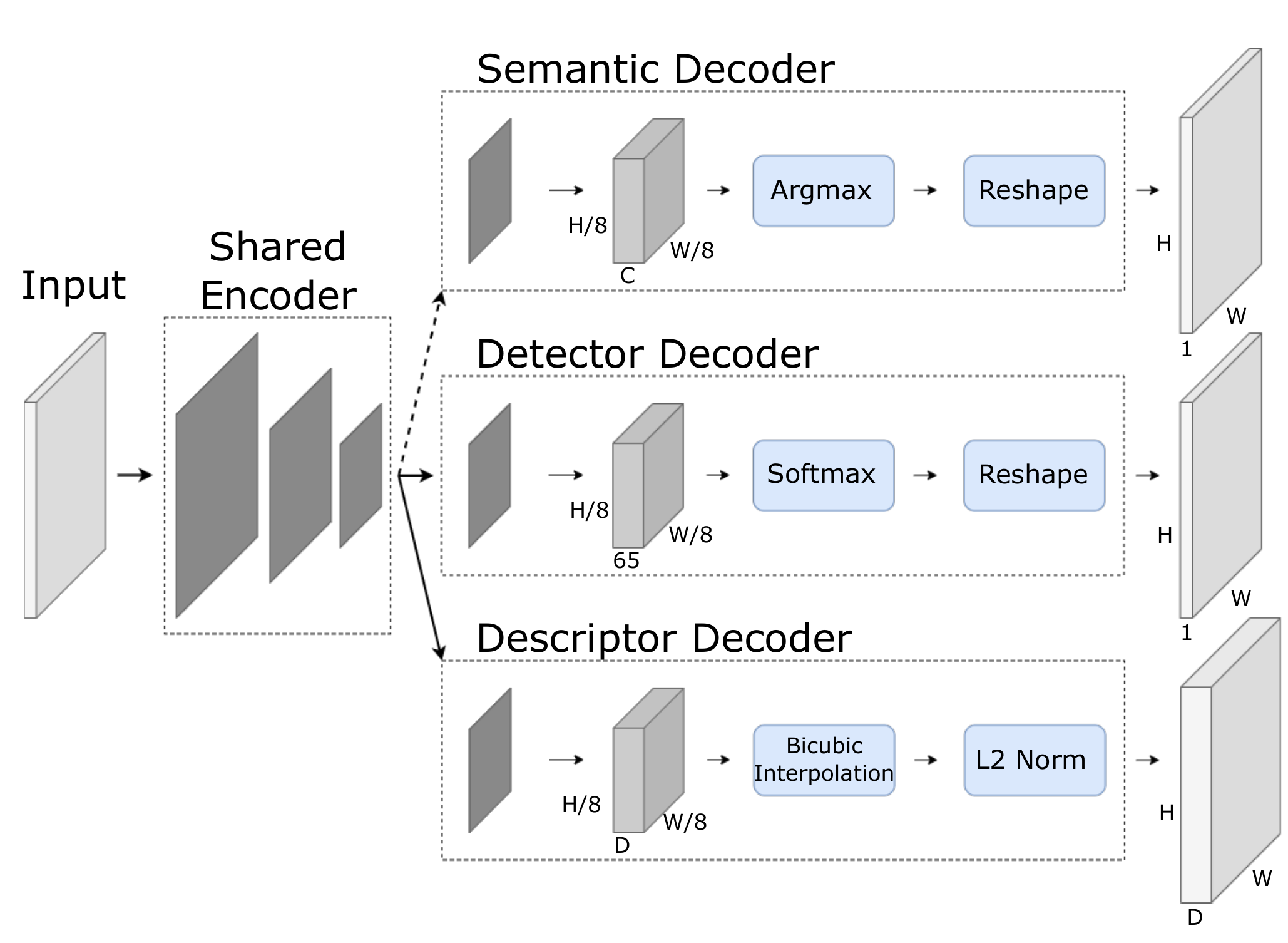}

    \caption{\textbf{Semantic SuperPoint Model}. The dashed line represents that the semantic decoder was only used in training and C is the number of classes.}
    \label{fig:ssp}
\end{figure}

This article proposes adding semantic information intrinsically to the keypoint and descriptor, which would help to detect meaningful pixels and create a better descriptor. In this way, the pose estimation method can consider this semantic information as an additional criterion in the matching process of the extracted features. To accomplish that, multi-task learning methods are essential to effectively train the model. A diagram of the proposed model can be seen in Figure~\ref{fig:ssp}.

In this work, we also evaluate the models trained with and without the semantic decoder on the HPatches \cite{hpatches_2017_cvpr} according to the metrics defined by DeTone et al. \cite{superpoint} and on the KITTI dataset \cite{kitti_seq} at the SLAM task using the ORB-SLAM2 \cite{orb-slam2} as the base framework. We show that the proposed Semantic SuperPoint (SSp) model improves the matching score metric and estimates better trajectories. For a 480×640 image, our model runs at \textbf{208 FPS} or \textbf{4.8 ms} per frame on an NVIDIA GTX 3090.

In summary, the key contributions of this paper are:

\begin{itemize}
    \item Evaluation of multi-task learning methods applied to SuperPoint-like models.
    \item A novel training paradigm for deep feature extraction.
    % \item A slight improvement on the SuperPoint model.
    \item A slight improvement in the Matching Score metric.
    
\end{itemize}

\section{RELATED WORK}

\textbf{Semantic SLAM}. Adding high-level semantic information to SLAM-related applications was previously reported in the literature on multiple occasions to improve methods based only on the intensities of the pixels. Examples of those projects are the use of high-level semantic information for loop closure detection with LiDAR 3D points to avoid local minimums \cite{semantic_scan}, detection of dynamic objects to calculate their estimated speed and take that into account for a more accurate pose estimation process \cite{dyna_SLAM}, segmentation of objects for better description considering a database of selected artifacts \cite{sSLAM4AUV}, and for a handmade semantic descriptor combined with a photometric one \cite{sem_vslam_health}.

All of those highly depend on the quality of the network prediction, therefore are sensible to reflections and unknown objects. Even though the same can be said about our model, the semantic information only helps to extract better features and is not used in the inference, so it is more robust to misclassification.

\textbf{Multi-task Learning}. The simplest way to address multiple tasks is to do a uniform sum of each loss, as presented in Equation \eqref{eq:mtl_sum}, in the specific case of $w_t = 1$ for every task $t$. This usually obtains the worst result because of the different nature of the tasks and scales of the loss functions used to minimize them. Consequently, the model can converge prioritizing one task over another. 

One way to avoid this is to compute a weighted sum of each loss according to a positive scalar $w_t$. This has a better result than the uniform sum. However, it does not achieve the same result when compared to multi-task learning methods, requiring a grid search for the weighting parameters, which is time-consuming and computationally costly. The mentioned approach tries to minimize the following loss objective:

\begin{equation}
    L_{total} = \sum_{t=1}^N w_t L_t \text{,}
    \label{eq:mtl_sum}
\end{equation}

\noindent where $w_t$ is the weight that scales the $t$-th task loss $L_t$, and $N$ is the number of tasks.

To avoid this undesired convergence behavior and the grid search, this work implements two multi-task learning methods proposed in \cite{MTL_kendall, MTL_ang}. 

The first method \cite{MTL_kendall} is based on the uncertainty modeling to weigh losses and derive a multi-task loss function by maximizing a Gaussian likelihood with homoscedastic uncertainty to classification and regression problems. 

Let $f^\mathbf{W}(\mathbf{x})$ be the output of a neural network with $\mathbf{W}$ weights and input $\mathbf{x}$ and $\mathbf{y}$ for the respective label. For regression problems, considering a normal distribution, the log-likelihood associated is:
\begin{equation}
    \text{log} \, p(\mathbf{y}|f^\mathbf{W}(\mathbf{x})) \propto -\frac{1}{2\sigma^2}||y -  f^\mathbf{W}(\mathbf{x})||^2 - \text{log}\, \sigma \text{,}
    \label{eq:mtl_kendall_reg}
\end{equation}

\noindent where $\sigma$ is an observation noise scalar. In the case of a classification problem:

\begin{equation}
    \text{log} \, (\mathbf{y}=c|f^\mathbf{W}(\mathbf{x}), \sigma) \propto -\frac{1}{\sigma^2}f^\mathbf{W}(\mathbf{x}) -\text{log}\, \sigma \text{,}
    \label{eq:mtl_kendall_class}
\end{equation}

\noindent where $c$ is the predicted class.

The second approach \cite{MTL_ang} consists in finding a common descent direction for the shared encoder. This is done by computing the gradient related to each task, normalizing them, and
searching for weighting coefficients that minimize the norm of the convex combination, obtaining the central direction.

To consider the gradient history, a period of $T$ iterations of the gradients’ norm of each task are tracked and, in case of some task diverging, the central dir. is pulled toward it as the tensioner's idea.
Nakamura et al. \cite{MTL_ang} shows that by doing this the shared encoder parameters are adjusted to equally benefit all tasks, besides decreasing the convergence time significantly.

For a more detailed explanation, please refer to the original article of each method.

\textbf{SuperPoint}. SuperPoint \cite{superpoint} is a deep feature extractor trained by self-supervision that can perform in real-time while executing both keypoint detection and descriptor decoders. 

For the detection decoder, a base detector MagicPoint is trained in the Synthetic Shapes dataset, witch renders simple 2D objects, such as quadrilaterals, triangles, lines, and ellipses. In the dataset, the keypoint label is created by marking certain junctions positions, end of segments, and ellipses center.

% To generate the pseudo-ground truth, a base detector called MagicPoint is trained in a Synthetic dataset generated on the fly which renders simple 2D objects, such as quadrilaterals, triangles, lines, and ellipses. In the dataset, the keypoint label is created by marking certain junctions positions, end of segments, and ellipses center.

The MagicPoint model is then used to extract the interest points in the selected dataset by performing a number $N_h$ of homographic adaptations and combining its output to generate a final heatmap, filtered by Non-Maximum Suppression (NMS) \cite{NMS}.

\begin{figure*}[htpb]
  \setbox9=\hbox{\centering\includegraphics[width=.32\linewidth]{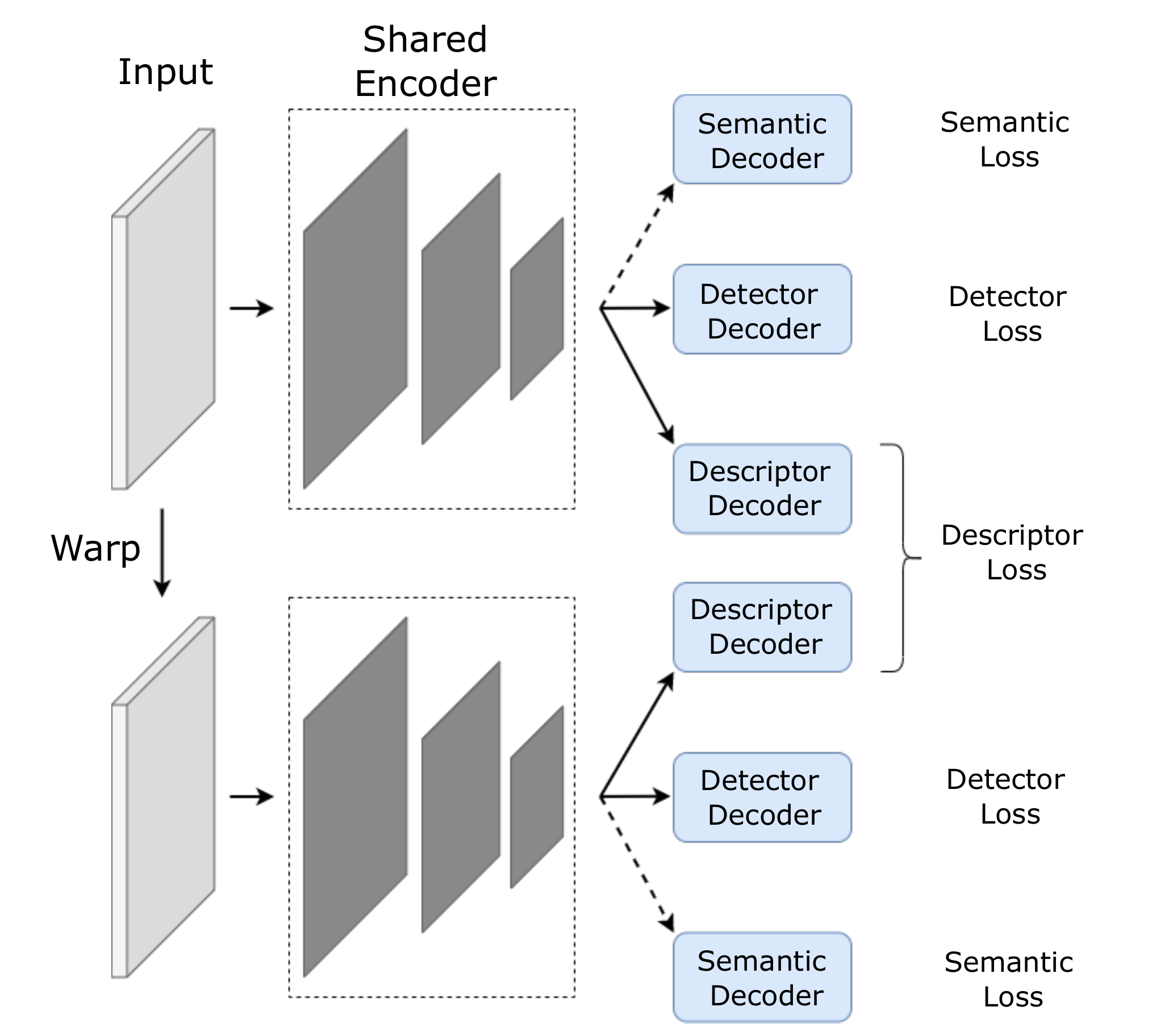}}% Capture tallest image in box 9
  \subcaptionbox{Training in Synthetic Shapes dataset}
    {\raisebox{\dimexpr\ht9/2-\height/2}{\includegraphics[width=.32\linewidth]{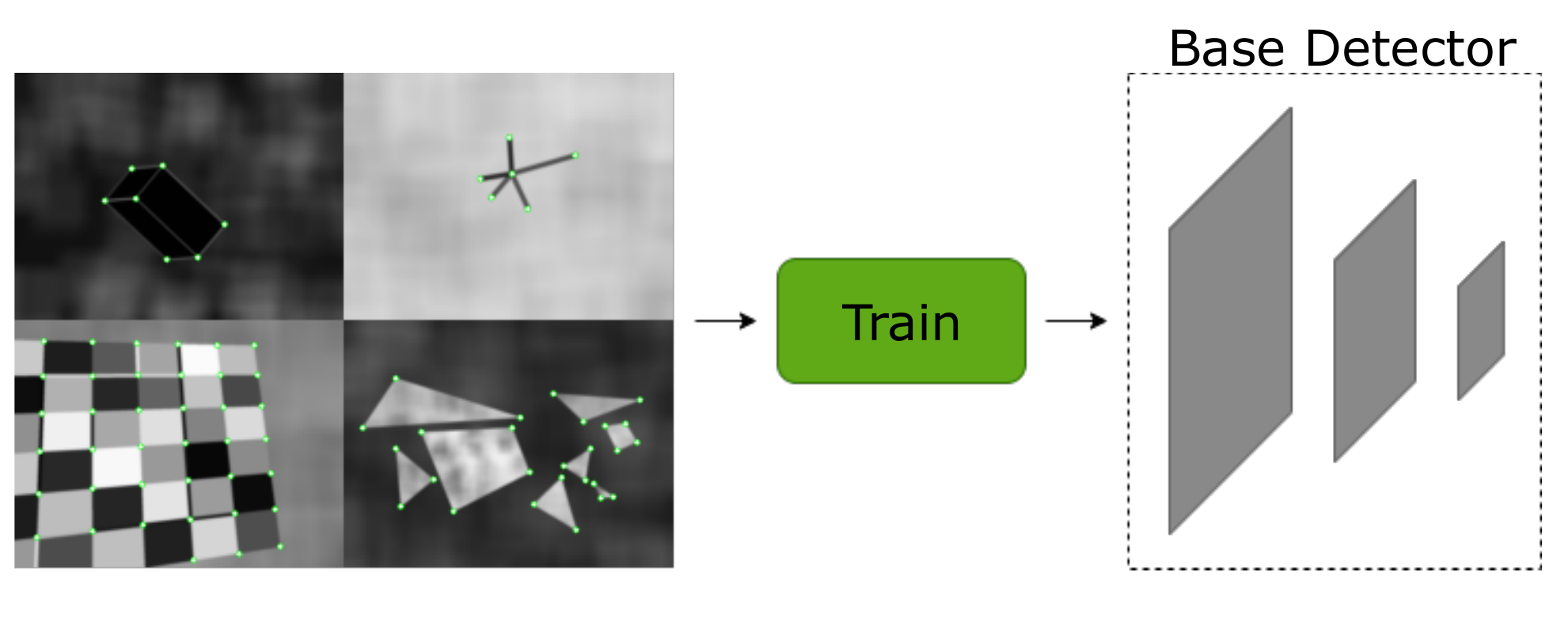}}} 
\rulesep
\hfill
  \subcaptionbox{Keypoint extraction}
    {\raisebox{\dimexpr\ht9/2-\height/2}{\includegraphics[width=.32\linewidth]{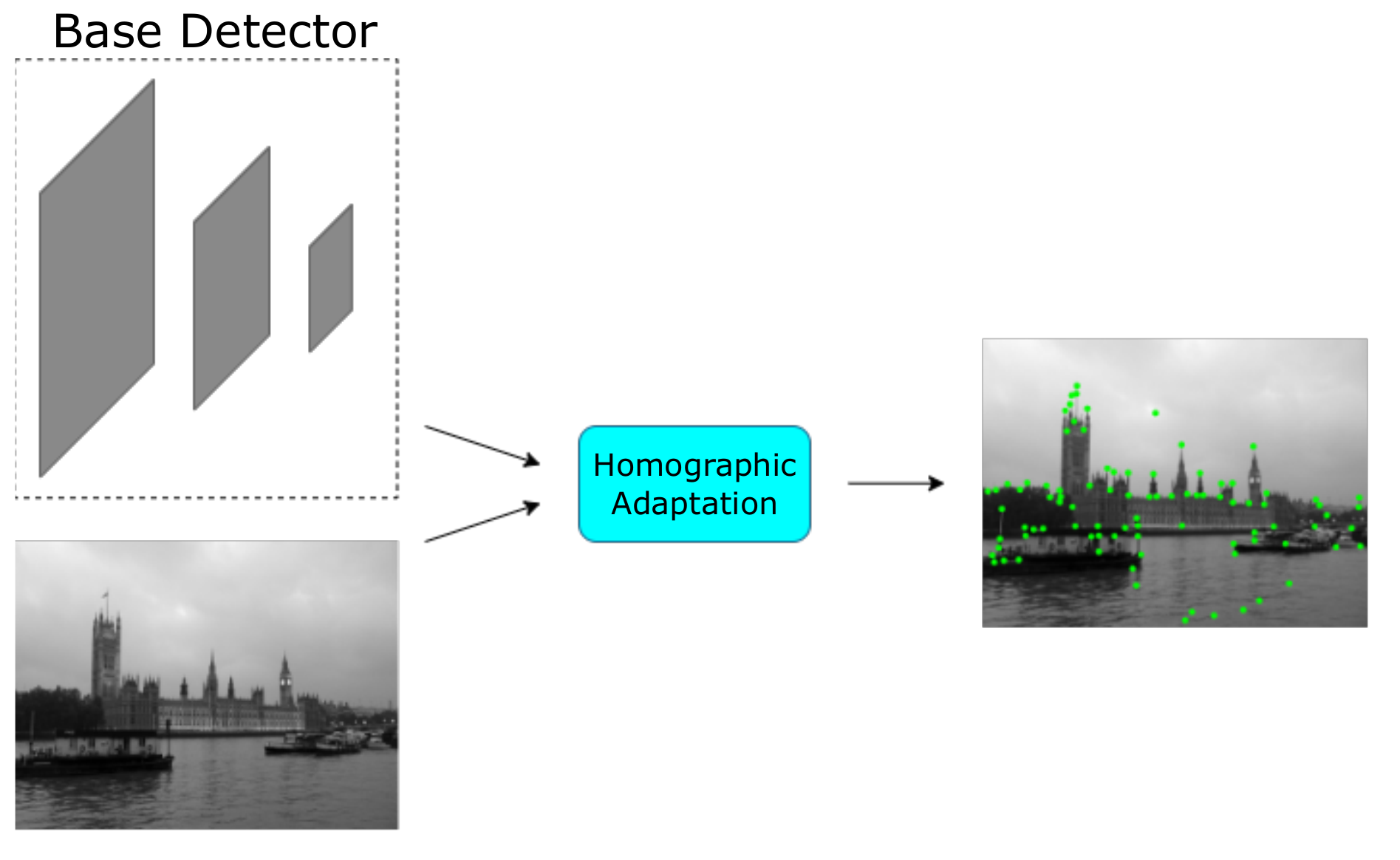}}}
\rulesep
\hfill
  \subcaptionbox{Joint training}{\includegraphics[width=.32\linewidth]{imgs/SP/SSP_joint_training-cropped.pdf}}

\caption{\textbf{Semantic SuperPoint training overview}. A base betector is trained in the Synthetic Shapes dataset and used to extract pseudo-ground truth interest points in unseen real images using $N_h$ homographic transformations. The joint training step uses the generated labels with the semantic label to train the SSp model. Adapted from DeTone et al. \cite{superpoint}.}
\label{fig:sp_pipeline}
\end{figure*}

The next step is joint training with both detector and descriptor decoders. A set with the original image and its warped version is passed through the model, outputting the detected interest points and the descriptors. Since the warp transformation is known, the same goes for the position relation between each pixel. Thus, pairs of matches and non-matches can be created and compared to the ground truth, illustrated in Figure \ref{fig:sp_pipeline}.

\section{SEMANTIC SUPERPOINT MODEL}
SSp is a fully convolutional neural network that receives a grayscale image as input and outputs a heatmap, descriptor, and semantic segmentation prediction. The architecture uses a single shared encoder and its output (intermediary features) passes through each decoder.

Due to the aggressive data augmentation process and the large variety of classes in the MS-COCO dataset \cite{MSCOCODATA}, used for training, the semantic prediction is not usable in practical application. Nonetheless, adding a semantic decoder for training improved the quality of the deep feature extractor. It can therefore be said that rather than being a model, SSp is more like a training method.

\subsection{Shared Encoder}
According to the open source implementation of SuperPoint (Sp) \cite{SP_open_git}, the encoder is based on the U-Net \cite{unet}, using \texttt{double\_conv} blocks,  which consists of $2\times$(conv2d-BN-ReLu) and three layers of max pooling to reduce the dimension between them. Therefore the output has a shape of $\frac{H}{8}\times \frac{W}{8}\times C_{enc}$. We adopt $C_{enc} = 256$ as the encoder output dimension. The same encoder model is used for Sp and SSp architectures.

\subsection{Detector Decoder}
Both the detector and descriptor decoders are the same as SuperPoint's original model. The detector has a shape of $\frac{H}{8}\times \frac{W}{8}\times 65$ to avoid upsampling operations and has one dummy dimension 'no keypoints'.
The loss function used was the binary cross entropy (BCE):
\begin{multline}
    L_{d} = \sum_{n=1}^{N} [y_n\log(\text{Softmax}(x_n))\\
            +(1-y_n)\log(1-\text{Softmax}(x_n))] \text{,}
    % L_{d} = -W_n[y_n\log(\text{Softmax}(x_n))\\
    %         {}+(1-y_n)\log(1-\text{Softmax}(x_n))]
\label{eq:det_loss}
\end{multline}
\noindent where N is the batch size.

\subsection{Descriptor Decoder}
For the descriptor decoder, also to avoid adding computational cost related to the upsampling, the output is shaped as $\frac{H}{8}\times \frac{W}{8} \times D$ and to extract the descriptor, it passes through a Bicubic Interpolation and is normalized.
To train the descriptor, it is necessary to relate two images and evaluate the matches and non-matches while balancing both influences. This was done by dividing the loss value by the number of positive ($n_p$) and negative ($n_n$) matches in the Hinge loss:
\begin{multline}
    L_{desc} = \underbrace{\frac{s}{n_p}\max(0,m_p-d^Td^{'})}_\text{$L_p$} \\
    + \underbrace{\frac{(1-s)}{n_n} \max(0,d^Td^{'} - m_n)}_\text{$L_n$}
    \text{,}
\label{eq:desc_loss}
\end{multline}
\noindent where $m_p = 1$, $m_n = 0.2$, $d$ and $d^{'}$ are the descriptor decoder outputs for the normal and warped image, respectively.

\subsection{Semantic Decoder}
The semantic decoder consists of a conv3x3-BN-ReLU block followed by a conv1x1. The cross-entropy (CE) loss was used for training this decoder:

\begin{equation}
    % L_{s} = -y_n\log ((x))
    L_{s} = -\sum_{n=1}^N \sum_{c=1}^{C} w_clog (\text{Softmax}(x_c))y_{n,c}.
    \label{eq:sem_loss}
\end{equation}

\section{MULTI-TASK LOSS} \label{sec:mtl}

In this section, we discuss the losses applied in the training process. In all cases, the semantic loss will be included to avoid rewriting the equations, but the non-semantic loss can be obtained by removing the $L_{s_i}$ terms. 

\subsection{Uniform loss} 
For comparison purposes, the model was also trained with the uniform sum loss function as follows: 

\begin{equation}
    L_{total} = L_{d_1} + L_{d_2} + \lambda L_{desc} + L_{s_1} + L_{s_2}.
    \label{eq:uni_loss_sp}
\end{equation}

The losses $L_{d_i}$ and $L_{s_i}$ are obtained from the interest point detector and the semantic decoder, respectively. The subscribed index 1 is for the normal image and 2 for the warped version.

In the original SuperPoint article \cite{superpoint}, they use the term $\lambda$ to weight the descriptor loss. In our case, the code was build on the unofficial implementation of Jau et al. \cite{SP_open_git}, and they use $\lambda=1$. Besides, the model was able to learn both tasks without the need for tuning the value $\lambda$.

\subsection{Uncertainty loss}

In the case of the uncertainty loss proposed by Kendall et al. \cite{MTL_kendall}, the detection loss can be directly related to the classification case, Equation \eqref{eq:mtl_kendall_class}, since the BCE loss behaves the same way as a CE loss for 2 classes. The same goes for the semantic decoder as it also uses CE.

The descriptor loss is more complex. To jointly calculate the terms of the hinge loss, the assumption that the losses are not correlated must be true. Even though this is not the case for most multi-task models, the approximation remains valid. 
The problem with the hinge loss is that the positive and negative terms are inversely proportional. The more the positive loss improves, the negative loss will not get better at the same rate as the model would create a bias towards positive matches and vice versa.

The descriptor loss was considered as two independent regression problems, each one with a related normal distribution. Considering the loss given by Equation \eqref{eq:desc_loss}, it is expected that the output $D_{out}=f^{\mathbf{W_{sh, desc}}}(\mathbf{x}_1)^Tf^{\mathbf{W_{sh, desc}}}(\mathbf{x}_2)$ is going to be grouped toward $m_p$ or $m_n$ for the positive matches ($s=1$) and the negative matches ($s=0$), respectively. So we can describe the distribution as:
\begin{equation}
    p(\mathbf{y}|D_{out}, s) = s \mathcal{N}_p (D_{out}, \sigma_p^2) + (1-s)\mathcal{N}_n (D_{out}, \sigma_n^2) \text{,}
    \label{eq:mtl_norm_desc}
\end{equation}
\noindent where $\mathcal{N}_i$ and $\sigma_i$ are the normal distribution and variance associated with each match type.

Expanding the normal distributions and approximating the loss terms, we obtain: 
\begin{equation}
    \text{log} \, p(\mathbf{y}|D_{out}) \propto -\frac{1}{2\sigma^2} \left(L_p + L_n\right) - \text{log} \, \sigma.
    \label{eq:mtl_norm_desc_vf}
\end{equation}

See Appendix \ref{ap:mtl_norm_desc} for more detail. The normal and warped loss terms are summed because they represent the same objective on the same weights, so the joint probability can be modeled as:

% This showed better results than the split version.

\begin{multline}
    p(\mathbf{y_{d_1},y_{d_2}, y_{desc}, y_{s_1}, y_{s_2}} | f^{\mathbf{W}}(\mathbf{x_1}), f^{\mathbf{W}}(\mathbf{x_2})) = \\ p(y_{d_1}, y_{d_2}|f^{\mathbf{W_{sh, d}}}(\mathbf{x_1}), f^{\mathbf{W}_{sh, d}}(\mathbf{x_2})) p(y_{desc} | D_{out})\\    
    p(\mathbf{y}_{s_1} = c_1, \mathbf{y}_{s_2} = c_2|f^{\mathbf{W_{sh, s}}}(\mathbf{x_1}), f^{\mathbf{W_{sh, s}}}(\mathbf{x_2})).
    \label{eq:loss_prob_t_v2}
\end{multline}

\noindent So the final equation for the uncertainty loss is given by:
\begin{multline}
    -log \, p(\mathbf{y_{d_1},y_{d_2}, y_{desc}, y_{s_1}, y_{s_2}} | f^{\mathbf{W}}(\mathbf{x_1}), f^{\mathbf{W}}(\mathbf{x_2})) \propto \\
     (L_{d_1}+L_{d_2}) \text{exp}(\eta_{d}) +\frac{L_{desc}}{2} \text{exp}(\eta_{desc})\\
    +(L_{s_1}+L_{s_2}) \text{exp}(\eta_{s}) + \eta_d + \frac{\eta_{desc}}{2} + \eta_s = L_{total}.
    \label{eq:mtl_uncertanty}
\end{multline}

In order to avoid numerical instability during training, we do the variable change: $\eta_i = 2\text{log} \, \sigma_i$. The initial values for $\eta_d, \, \eta_{desc},\, \eta_s$ were $1.0, \, 2.0, \, 1.0$ , respectively.

\subsection{Central dir. + tensor}
To apply the Central dir. + tensor method the normal and warped loss were summed to simplify the common gradient descent calculation and the descriptor loss was treated as a single task.  

The $\alpha$ term, which is used to regulate the tension factor’s sensitivity, was set to $0.3$, as indicated by the original authors \cite{MTL_ang}.

\section{DATA PREPARATION}

The MS-COCO 2017 dataset consists of 123k 640×480 RGB images split into 118k images for training and 5k for validation, with 133 segmented classes. This large amount and variability of data can provide the model with the necessary generalization capability to perform in several conditions while being similar to the 2014 one, used to train the original SuperPoint model.

\subsection{Keypoint extraction}

In order to create the pseudo-ground truth labels for the keypoint extractor, a pretrained MagicPoint model in the Synthetic Shapes dataset from Jau et al. \cite{SP_open_git} was applied to the MS-COCO dataset. It was used $N_h = 100$ as the number of homograph transformations, resized to a resolution of $240\times320$ to decrease training time.

\section{TRAINING}

All training and evaluation processes were done in a machine with an AMD Ryzen 9 5950X 16-Core Processor and an NVIDIA GTX 3090 24GB using PyTorch \cite{PyTorch}.

For data augmentation, multiple transformations, such as Gaussian noise, scale, rotation, translation, and others were applied to improve the network's robustness to illumination and viewpoint changes. The descriptor dimension is $D=256$ for all experiments.

\subsection{Sp \& SSp training}

The Sp and SSp models were trained using the Adam optimizer \cite{Adam} for 200k iterations with batch size 16. We tested with a fixed learning rate as in the original article, and with a starting value of $0.0025$ with polynomial decay and an end value of $0.001$. The first option worked better for the non-semantic version and the second one was better for the SSp model. 

We also varied the multi-task loss used for each model, as described in Section \ref{sec:mtl}. Empirically, we found that the central dir. $+$ tensor method took too long to converge. So we used the model saved at the 100K$^{th}$ iteration trained by the uncertainty loss as the starting point, and trained for more 100K iterations using the central dir. $+$ tensor optimization method with the same learning rate parameters.

In every variation, the models were saved every interval of 5K iterations and evaluated as described in Section \ref{sec:eval}.

To rank each model, we adopt the matching score as the main metric, since it is the percentage of inliers. Hence, it represents an overall performance of the keypoint detection and descriptor generation. Furthermore, it is the most relatable to SLAM methods.

\section{EVALUATION} \label{sec:eval}

\subsection{HPatches}

The HPatches dataset consists of 116 scenes each one with six different images. The first 57 scenes are subject to illumination changes and in the other 59 scenes, viewpoint changes.

The models were evaluated according to the metrics: repeatability (Rep.), localization error (MLE), homography estimation ($e=\{1, 3 , 5\}$), nearest neighbor mean average precision  (NN mAP), and matching score (M.S.). Each metric is described in detail by DeTone et al. \cite{superpoint}.

Considering the abbreviations \texttt{unc} for the uncertainty loss, \texttt{ct} for central dir. $+$ tensor and \texttt{uni} for the uniform loss, the results of each model are presented in Table \ref{tab:hpatches_metric}.

\begin{table}[ht]
\caption{Evaluation on HPatches dataset}
\label{tab:hpatches_metric}
\begin{center}
\resizebox{\linewidth}{!}{
{\renewcommand{\arraystretch}{1.2} %<- modify value to suit your needs

\begin{tabular}{|c|c|c|c|c|c|c|c|}
\hline

\multicolumn{1}{|c|}{\multirow{2}{3cm}{\centering Feature extractor}} & \multicolumn{3}{c|}{Homography Estimation} & \multicolumn{2}{c|}{Detector metrics} & \multicolumn{2}{c|}{Descriptor metrics}\\ \cline{2-8}

 & $\epsilon = 1$ & $\epsilon = 3$ & $\epsilon = 5$ & Rep. & MLE & NN mAP. & M.S.\\
  
  % & $\epsilon = 1$  \textuparrow & $\epsilon = 3$  \textuparrow& $\epsilon = 5$  \textuparrow & Rep.  \textuparrow & MLE  \textdownarrow& NN mAP.  \textuparrow & M.S. \textuparrow\\
% \cline{2-8}
\hline

ORB$^{\mathrm{a}}$ & .150 & .395 & .538 & \textbf{.641} & 1.157 & .735 & .266 \\
SIFT$^{\mathrm{a}}$ & .424 & .676 & .759 & .495 & \textbf{0.833} & .694 & .313 \\
LIFT$^{\mathrm{a}}$ & .284 & .598 & .717 & .449 & 1.102 & .664 & .315 \\
\hline
Sp + uni (baseline) & .476 & .748 & .817 & .599 & 1.017 & \textbf{.864} & .519 \\

Sp + unc & .460 & .745 & .812 & .599 & 1.010 & \textbf{.864} & .520\\

Sp + ct & .493 & .753 & .812 & .602 & 1.001 & .862 & .519\\
\hline

SSp + uni (\textbf{ours})  & .398 & .710 & .797 & .584 & 1.052 & .843 & .506\\
SSp + unc (\textbf{ours})  & .450 & .745 & .816 & .598 & 1.005 & \textbf{.864} & .522\\
SSp + ct (\textbf{ours})  & .466 & .762 & .805 & .598 & 0.999 & .858 & .519\\ \hline
Pretrained model \cite{superpoint} & \textbf{.497} & \textbf{.766} & \textbf{.840} &	.610 & 1.086 & .843 & \textbf{.540} \\
\hline
% \multicolumn{7}{l}{$^{\mathrm{a}}$Original results reported by DeTone et al. \cite{superpoint}.}
\multicolumn{7}{l}{$^{\mathrm{a}}$Original results reported by DeTone et al. (2017).}
\end{tabular}}}

% TODO: n sei se essa é a maneira certa
\caption*{All metrics are higher is better, except MLE.}

\end{center}
% \footnotesize{$^1$ Original results reported by DeTone et al. \cite{superpoint}.}
\end{table}

Analyzing the results of the SSp model, the uniform loss had the worst result
compared to the others SuperPoint models. The uncertainty and central dir. methods improved significantly the initial result obtained using the uniform loss for the SSp model and had a very similar performance in the case of the Sp model.

Considering the matching score, the SSp model had the best result with a slight improvement of $\mathbf{+}$\textbf{0.2\%} over the baseline Sp. As for the Homography Estimation it had a better result than the Sp + unc for $\epsilon = 5$ while balancing the detector metrics.

Thus, adding semantic information can help deep feature extraction methods, but still needs some improvement to increase all metrics.

It was not possible to replicate the result obtained from the Magic Leaps pretrained model \cite{superpoint} since the official implementation is not public available as well its specifics. For this reason, the results were not directly compared with the pretrained model, instead we considered the Sp + uni as baseline for comparison.

\subsection{KITTI}
The best version of each model was evaluated in the KITTI dataset according to the implementation SuperPoint\_SLAM \cite{ORB_SLAM2_SP}. Each sequence was executed 10 times and the Absolute Pose Error (APE) and Relative Pose Error (RPE) metrics were extracted with the open source lib \texttt{evo} \cite{evo}. The mean values and the standard variation together with the p-values are exhibited in Table \ref{tab:KITTI_metrics}.  

\begin{table*}[ht]

\sisetup{
  table-align-uncertainty=true,
  separate-uncertainty=true,
}
%% local redefinitions
\renewrobustcmd{\bfseries}{\fontseries{b}\selectfont}
\renewrobustcmd{\boldmath}{}

\centering

\caption{Mean APE and RPE in the KITTI dataset for each model}
% \begin{tabular}{|c||ccc||ccc|}
% \begin{tabular}{|cS[table-format=3.3(3),detect-weight,mode=text]S[table-format=3.3(3),detect-weight,mode=text]c|S[table-format=3.3(3),detect-weight,mode=text]S[table-format=3.3(3),detect-weight,mode=text]c|}

\begin{tabular}{|c||S[table-format=3.3,detect-weight,mode=text] @{${}\pm{}$} S[table-format=3.3,detect-weight,mode=text]|S[table-format=3.3,detect-weight,mode=text] @{${}\pm{}$} S[table-format=3.3,detect-weight,mode=text]|c||S[table-format=1.3,detect-weight,mode=text] @{${}\pm{}$} S[table-format=1.3,detect-weight,mode=text]|S[table-format=1.3,detect-weight,mode=text] @{${}\pm{}$} S[table-format=1.3,detect-weight,mode=text]|c|}

\hline

\multicolumn{1}{|c||}{\multirow{2}{3cm}{\centering Sequence}} & \multicolumn{5}{c||}{\centering ATE} & \multicolumn{5}{c|}{RPE} \\  \cline{2-11}

& \multicolumn{2}{c|}{\centering Sp} & \multicolumn{2}{c|}{\centering SSp} & p-value & \multicolumn{2}{c|}{\centering Sp} & \multicolumn{2}{c|}{\centering SSp} & p-value\\
\hline

00 &  6.767 & 1.112 & \bfseries 6.651 & \bfseries 0.614 & 0.705 & \bfseries 0.124 & \bfseries 0.028 & 0.130 & 0.032 & 0.804 \\
01 &  286.709 & 140.875 & \bfseries 209.985 & \bfseries 130.585 & 0.131 & 4.779 & 1.910 & \bfseries 3.991 & \bfseries 2.183 & 0.174 \\
02 &  22.459 & 2.428 & \bfseries 22.314 & \bfseries 2.988 & 0.821 & \bfseries 0.098 & \bfseries 0.008 & \bfseries 0.098 & \bfseries 0.009 & 0.705 \\
03 &  \bfseries 1.319 & \bfseries 0.142 & 1.630 & 0.262 & 0.003 & \bfseries 0.040 & \bfseries 0.002 & 0.046 & 0.004 & 0.007 \\
04 &  \bfseries 0.840 & \bfseries 0.135 & 0.905 & 0.261 & 0.940 & \bfseries 0.034 & \bfseries 0.004 & 0.039 & 0.008 & 0.406 \\
05 &  \bfseries 5.803 & \bfseries 1.233 & 6.315 & 2.931 & 0.821 & \bfseries 0.172 & \bfseries 0.046 & 0.199 & 0.101 & 1.000 \\
06 &  11.953 & 0.939 & \bfseries 11.833 & \bfseries 0.415 & 0.406 & \bfseries 0.240 & \bfseries 0.048 & 0.267 & 0.075 & 0.326 \\
07 &  3.388 & 2.947 & \bfseries 2.124 & \bfseries 0.516 & 1.000 & 0.108 & 0.071 & \bfseries 0.084 & \bfseries 0.010 & 0.884 \\
08 &  31.324 & 1.736 & \bfseries 26.700 & \bfseries 1.013 & 0.000 & 0.273 & 0.012 & \bfseries 0.235 & \bfseries 0.009 & 0.000 \\
09 &  35.945 & 2.140 & \bfseries 31.788 & \bfseries 1.595 & 0.001 & 0.172 & 0.013 & \bfseries 0.164 & \bfseries 0.010 & 0.226 \\
10 &  5.515 & 0.437 & \bfseries 4.953 & \bfseries 0.316 & 0.023 & 0.064 & 0.003 &  \bfseries 0.060 & \bfseries 0.003 & 0.019 \\

\hline
\end{tabular}
\label{tab:KITTI_metrics}
\end{table*}

Note that the SLAM application used here is only for comparison purposes as the matching time adds to much computational cost, being responsible for most of it \cite{SP_time}. To deal with these issues, approaches similar to GCNv2 \cite{SP_time} can be used.

\section{CONCLUSIONS}

We have shown that adding semantic information in a shared encoder based architecture in combination with multi-task learning methods can improve deep feature extraction methods and, in this case, without increasing computational cost in the inference step. 

Future works can tune the intensity of the data augmentation process and the complexity of the training dataset to balance generalization capability and semantic learning or train the feature extractor using another approach that does not rely on intensive data augmentation.

Additionally, if achieved a satisfying semantic prediction, it is possible to combine the high-level semantic information into the already existing semantic SLAM methods for further improvements.

\addtolength{\textheight}{-2cm}   % This command serves to balance the column lengths
\section*{APPENDIX}

\subsection{Descriptor distribution} \label{ap:mtl_norm_desc}

The descriptor distribution is initially defined as two normal distributions as in Equation \eqref{eq:mtl_norm_desc}, copied bellow for convenience:
\begin{equation}
    p(\mathbf{y}_{desc}|D_{out}, s) = s \mathcal{N}_p (D_{out}, \sigma_p^2) + (1-s)\mathcal{N}_n (D_{out}, \sigma_n^2).
    \label{eq:mtl_norm_desc_copy}
\end{equation}

\noindent Expanding the norm distribution we have:
\begin{multline}
    p(\mathbf{y}_{desc}|D_{out}, s) = \frac{s}{\sigma_p \sqrt{2\pi}} \exp{\left[\frac{-1}{2} \left(\frac{\mathbf{y}_{desc}-D_{out}}{\sigma_p} \right)^2 \right]} \\
    + \frac{1-s}{\sigma_n \sqrt{2\pi}} \exp{\left[\frac{-1}{2} \left(\frac{\mathbf{y}_{desc}-D_{out}}{\sigma_n} \right)^2 \right]}. 
    \label{eq:mtl_norm_desc_extended}
\end{multline}

Considering the same variance for both losses and substituting the term $(y-D)^2$ for the respective loss we have:\cite{orb-slam2}
\begin{equation}
    p(\mathbf{y}_{desc}|D_{out}) = \frac{1}{\sigma\sqrt{2\pi}} \left[ \exp\left( { \frac{-L_p}{2\sigma^2} }\right) 
    + \exp{\left(\frac{-L_n}{2\sigma^2} \right)}\right].
\end{equation}

The $s$ term is no longer needed because each loss is related to each match case. Since we are not balancing the losses between each other, we define the likelihood as:
\begin{equation}
    p(\mathbf{y}_{desc}|D_{out}) \propto \frac{1}{\sigma \sqrt{2\pi}} \exp\left( { \frac{-(L_p + L_n)}{2\sigma^2} }\right).
\end{equation}

\noindent So the log likelihood is:
\begin{equation}
    \text{log} \, p(\mathbf{y}_{desc}|D_{out}) \propto -\frac{1}{2\sigma^2} \left(L_p + L_n\right) - \text{log} \, \sigma.
    \label{eq:mtl_norm_desc_vf_ap}
\end{equation}

\bibliographystyle{plain}

\bibliography{ref}

\end{document}